\PassOptionsToPackage{unicode}{hyperref}
\PassOptionsToPackage{hyphens}{url}
\PassOptionsToPackage{dvipsnames,svgnames,x11names}{xcolor}
\documentclass[
  12pt]{article}

\usepackage{amsmath,amssymb}
\usepackage{iftex}
\ifPDFTeX
  \usepackage[T1]{fontenc}
  \usepackage[utf8]{inputenc}
  \usepackage{textcomp} 
\else 
  \usepackage{unicode-math}
  \defaultfontfeatures{Scale=MatchLowercase}
  \defaultfontfeatures[\rmfamily]{Ligatures=TeX,Scale=1}
\fi
\usepackage{lmodern}
\ifPDFTeX\else  
\fi
\IfFileExists{upquote.sty}{\usepackage{upquote}}{}
\IfFileExists{microtype.sty}{
  \usepackage[]{microtype}
  \UseMicrotypeSet[protrusion]{basicmath} 
}{}
\makeatletter
\@ifundefined{KOMAClassName}{
  \IfFileExists{parskip.sty}{%
    \usepackage{parskip}
  }{
    \setlength{\parindent}{0pt}
    \setlength{\parskip}{6pt plus 2pt minus 1pt}}
}{
  \KOMAoptions{parskip=half}}
\makeatother
\usepackage{xcolor}
\makeatletter
\ifx\paragraph\undefined\else
  \let\oldparagraph\paragraph
  \renewcommand{\paragraph}{
    \@ifstar
      \xxxParagraphStar
      \xxxParagraphNoStar
  }
  \newcommand{\xxxParagraphStar}[1]{\oldparagraph*{#1}\mbox{}}
  \newcommand{\xxxParagraphNoStar}[1]{\oldparagraph{#1}\mbox{}}
\fi
\ifx\subparagraph\undefined\else
  \let\oldsubparagraph\subparagraph
  \renewcommand{\subparagraph}{
    \@ifstar
      \xxxSubParagraphStar
      \xxxSubParagraphNoStar
  }
  \newcommand{\xxxSubParagraphStar}[1]{\oldsubparagraph*{#1}\mbox{}}
  \newcommand{\xxxSubParagraphNoStar}[1]{\oldsubparagraph{#1}\mbox{}}
\fi
\makeatother

\usepackage{longtable,booktabs,array}
\usepackage{calc} 
\usepackage{etoolbox}
\makeatletter
\patchcmd\longtable{\par}{\if@noskipsec\mbox{}\fi\par}{}{}
\makeatother
\IfFileExists{footnotehyper.sty}{\usepackage{footnotehyper}}{\usepackage{footnote}}
\makesavenoteenv{longtable}
\usepackage{graphicx}
\makeatletter
\def\maxwidth{\ifdim\Gin@nat@width>\linewidth\linewidth\else\Gin@nat@width\fi}
\def\maxheight{\ifdim\Gin@nat@height>\textheight\textheight\else\Gin@nat@height\fi}
\makeatother
\setkeys{Gin}{width=\maxwidth,height=\maxheight,keepaspectratio}
\makeatletter
\def\fps@figure{htbp}
\makeatother

\makeatletter
\@ifpackageloaded{caption}{}{\usepackage{caption}}
\AtBeginDocument{%
\ifdefined\contentsname
  \renewcommand*\contentsname{Table of contents}
\else
  \newcommand\contentsname{Table of contents}
\fi
\ifdefined\listfigurename
  \renewcommand*\listfigurename{List of Figures}
\else
  \newcommand\listfigurename{List of Figures}
\fi
\ifdefined\listtablename
  \renewcommand*\listtablename{List of Tables}
\else
  \newcommand\listtablename{List of Tables}
\fi
\ifdefined\figurename
  \renewcommand*\figurename{Figure}
\else
  \newcommand\figurename{Figure}
\fi
\ifdefined\tablename
  \renewcommand*\tablename{Table}
\else
  \newcommand\tablename{Table}
\fi
}
\@ifpackageloaded{float}{}{\usepackage{float}}
\floatstyle{ruled}
\@ifundefined{c@chapter}{\newfloat{codelisting}{h}{lop}}{\newfloat{codelisting}{h}{lop}[chapter]}
\floatname{codelisting}{Listing}

\makeatother
\makeatletter
\@ifpackageloaded{caption}{}{\usepackage{caption}}
\@ifpackageloaded{subcaption}{}{\usepackage{subcaption}}
\makeatother

\ifLuaTeX
  \usepackage{selnolig}  
\fi
\usepackage[]{natbib}
\usepackage{bookmark}

\IfFileExists{xurl.sty}{\usepackage{xurl}}{} 
\hypersetup{
  pdftitle={Title},
  pdfauthor={Author 1; Author 2},
  pdfkeywords={3 to 6 keywords, that do not appear in the title},
  colorlinks=true,
  linkcolor={blue},
  filecolor={Maroon},
  citecolor={Blue},
  urlcolor={Blue},
  pdfcreator={LaTeX via pandoc}}

\usepackage{algorithm}
\usepackage{algpseudocode}
\usepackage{tabularx}
\usepackage{multirow, mathtools}
\usepackage{tikz}
\usepackage{setspace}

\usepackage{amsmath,amsfonts,bm,amssymb,amsthm}
\theoremstyle{plain}
\newtheorem{theorem}{Theorem}[section]
\newtheorem{proposition}[theorem]{Proposition}

\theoremstyle{definition}

\theoremstyle{remark}

\def\eqref#1{equation~\ref{#1}}

\def\1{\bm{1}}

\def\vtheta{{\bm{\theta}}}
\def\vdelta{{\bm{\delta}}}

\def\va{{\bm{a}}}

\def\ve{{\bm{e}}}

\def\vu{{\bm{u}}}
\def\vv{{\bm{v}}}
\def\vw{{\bm{w}}}
\def\vx{{\bm{x}}}

\def\vz{{\bm{z}}}
\def\vbeta{{\bm{\beta}}}

\def\mA{{\bm{A}}}

\def\mX{{\bm{X}}}

\DeclareMathAlphabet{\mathsfit}{\encodingdefault}{\sfdefault}{m}{sl}
\SetMathAlphabet{\mathsfit}{bold}{\encodingdefault}{\sfdefault}{bx}{n}

\def\gB{{\mathcal{B}}}

\def\gD{{\mathcal{D}}}

\def\gP{{\mathcal{P}}}
\def\gQ{{\mathcal{Q}}}
\def\gR{{\mathcal{R}}}
\def\gS{{\mathcal{S}}}

\def\gX{{\mathcal{X}}}
\def\gY{{\mathcal{Y}}}

\newcommand{\E}{\mathbb{E}}

\newcommand{\R}{\mathbb{R}}

\DeclareMathOperator*{\argmax}{arg\,max}

\DeclareMathOperator{\sign}{sign}

\makeatletter
\newcommand{\multiline}[1]{%
  \begin{tabularx}{\dimexpr\linewidth-\ALG@thistlm}[t]{@{}X@{}}
    #1
  \end{tabularx}
}
\makeatother

\newcommand{\anon}{1}

\begin{document}

\def\spacingset#1{\renewcommand{\baselinestretch}%
{#1}\small\normalsize} \spacingset{1}


\if1\anon
{
  \title{\bf Improving Clean Accuracy via a Tangent-Space Perspective on Adversarial Training}
  \author{Bongsoo Yi\thanks{
    The authors gratefully acknowledge support from NSF grants DMS-2152289, DMS-2134107, DMS-2401297, and Cisco Gift Fund.}\hspace{.2cm}\\
    Department of Statistics and Operations Research, \\ University of North Carolina at Chapel Hill\\
    Rongjie Lai \\
    Department of Mathematics, Purdue University \\
    Yao Li \\
    Department of Statistics and Operations Research, \\
    University of North Carolina at Chapel Hill}
  \maketitle
} \fi

\if0\anon
{
  \bigskip
  \bigskip
  \bigskip
  \title{\bf Improving Clean Accuracy via a Tangent-Space Perspective on Adversarial Training}
  \medskip
  \maketitle
} \fi

\bigskip
\begin{abstract}
Adversarial training has proven effective in improving the robustness of deep neural networks against adversarial attacks. However, this enhanced robustness often comes at the cost of a substantial drop in accuracy on clean data. In this paper, we address this limitation by introducing Tangent Direction Guided Adversarial Training (TART), a novel method that enhances clean accuracy by exploiting the geometry of the data manifold. We argue that adversarial examples with large components in the normal direction can overly distort the decision boundary and degrade clean accuracy. TART addresses this issue by estimating the tangent direction of adversarial examples and adaptively modulating the perturbation bound based on the norm of their tangential component. To the best of our knowledge, TART is the first adversarial defense framework that explicitly incorporates the concept of tangent space and direction into adversarial training. Extensive experiments on both synthetic and benchmark datasets demonstrate that TART consistently improves clean accuracy while maintaining robustness against adversarial attacks.
\end{abstract}

\noindent%

\vfill

\newpage
\spacingset{1.8} 

\section{Introduction}

Deep neural networks (DNNs) have achieved remarkable success in diverse domains such as computer vision~\citep{Krizhevsky2012ImageNetCW, He2015DeepRL, yi2023mlp, yi2024cnn}, natural language processing~\citep{otter2020survey}, recommendation systems~\citep{He2017NeuralCF}, and reinforcement learning~\citep{Mnih2015HumanlevelCT}. However, they remain highly vulnerable to adversarial examples, which are inputs perturbed by imperceptible, malicious noise intentionally crafted to induce incorrect predictions~\citep{Szegedy2013IntriguingPO, Goodfellow2015ExplainingAH, Nguyen2014DeepNN, jin2020bert}. This vulnerability poses critical security threats, particularly in safety-sensitive applications such as autonomous driving~\citep{Chen2015DeepDrivingLA, rossolini2023real} and medical diagnosis~\citep{Ma2019UnderstandingAA, Finlayson2019AdversarialAO}, where even a minor error can lead to catastrophic consequences.

To address these issues, developing effective defense techniques against adversarial attacks has become a central focus in machine learning research. One of the most effective methods is \textit{standard adversarial training}~\citep{madry2018towards}, which improves model robustness by incorporating adversarial examples into the training process. Following its initial success, a variety of extensions have been proposed~\citep{Gowal2020UncoveringTL, Zhang2019TRADES, Carmon2019UnlabeledDI, Xie2018FeatureDF, Wong2020Fast, Zhang2020FAT}. For a more comprehensive review of related literature, please refer to Appendix~\ref{app:related_work}. Despite these efforts, adversarial training has consistently shown an inherent trade-off: improving robustness often leads to a notable drop in clean accuracy~\citep{Zhang2019TRADES, Yang2020ACL, Tsipras2018RobustnessMB}. The primary goal of our study is to recover clean accuracy in adversarially trained models, without sacrificing their robustness. We tackle this problem by leveraging the geometric structure of the data manifold.

While many recent studies have focused on the connection between margin and adversarial learning in the ambient space~\citep{Ding2020MMA,zhang2021GAIRAT, Wang2021MAIL}, a growing body of work has begun to explore the role of data manifold geometry in adversarial defense~\citep{lin2020dual, xiao2025understanding, li2025enhancing}. The study of adversarial examples from a manifold perspective is motivated by two key observations: first, image data lies on a low-dimensional manifold embedded in a high-dimensional input space~\citep{Levina2004MaximumLE, pope2021the}; and second, adversarial examples are often located off this manifold~\citep{Jha2018DetectingAE, Tanay2016ABT, Li2021TowardsRO}. Despite this growing interest in manifold-based approaches, to the best of our knowledge, no prior work has directly leveraged the tangent space of the data manifold to guide adversarial training.

\begin{figure*}[t]
\centering
\includegraphics[width=\textwidth, trim={2cm 2.4cm 2cm 3.1cm},clip]{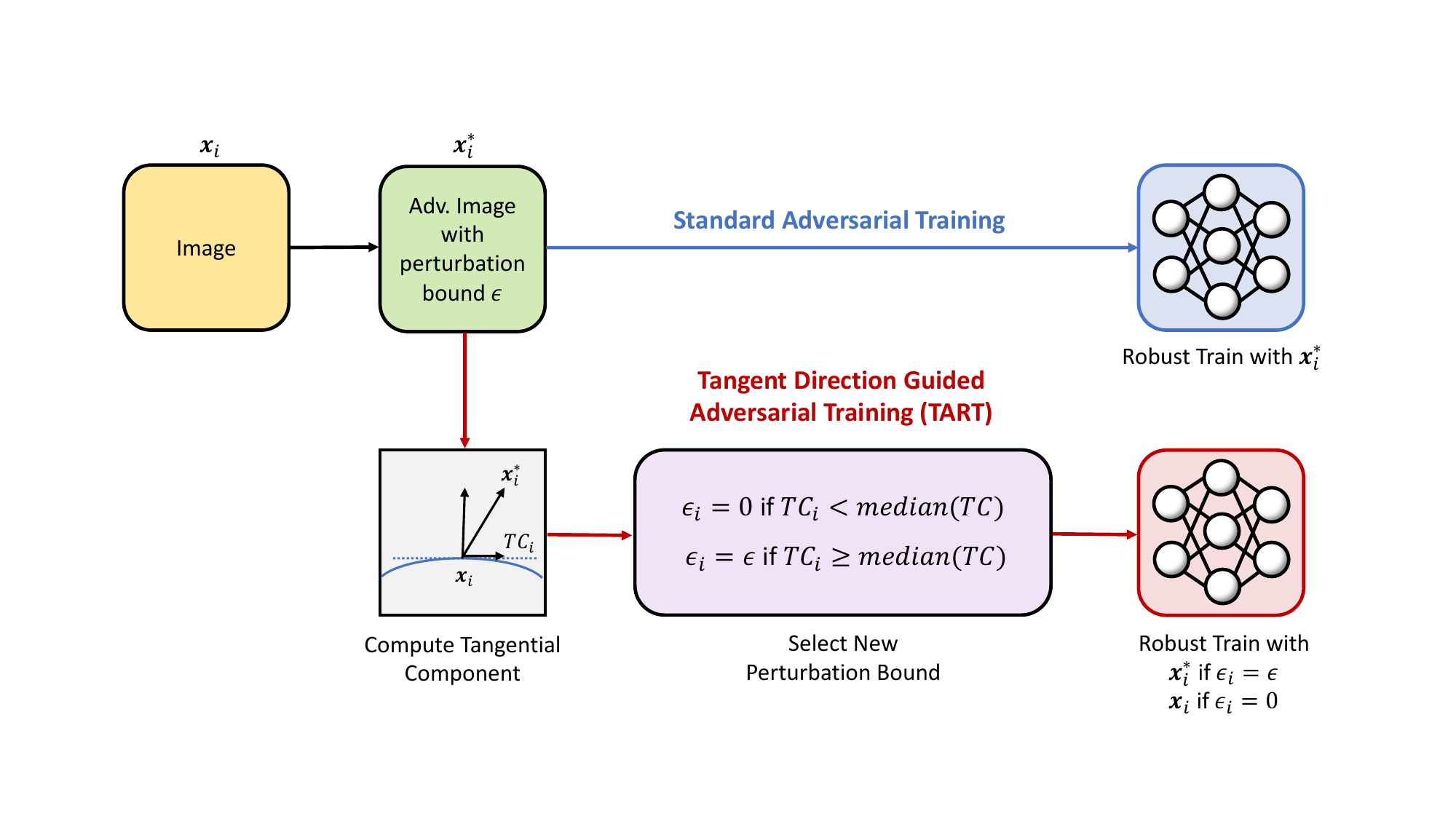}
\caption{Overview of TART and comparison with standard adversarial training. Given a training image $\vx_i$, we first generate an adversarial example $\vx_i^*$ with a fixed perturbation bound $\epsilon$. Standard adversarial training trains a robust model using these examples $\vx_i^*$. 
In contrast, TART first estimates and stores the tangent space of each training image offline using a pre-trained autoencoder and principal component analysis (PCA). Then, based on the stored tangent space information, TART computes the tangential component of $\vx_i^*$. 
TART finally uses $\vx_i^*$ for robust training if its tangential component falls within the upper 50\%, and employs $\vx_i$ if the tangential component is in the lower 50\%.}
\label{fig:TART}
\end{figure*}

\begin{figure}[t]
\centering
\begin{subfigure}{0.4\textwidth}
\includegraphics[width=\textwidth]{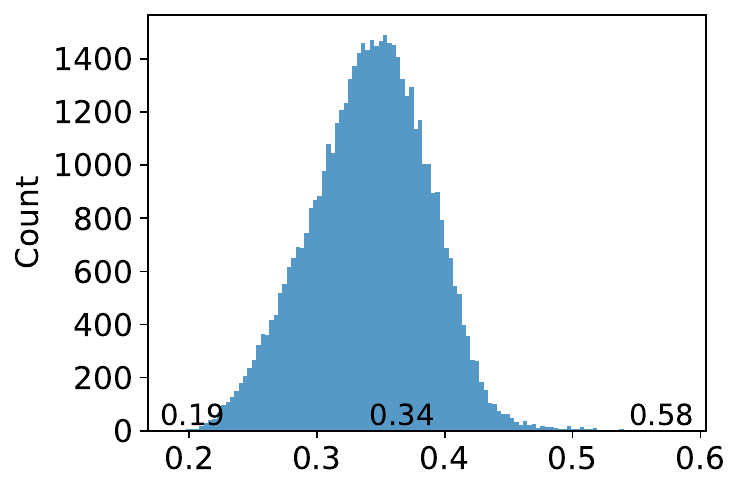}
\caption{Tangential Component }
\label{fig:tan_distribution}
\end{subfigure}
\begin{subfigure}{0.4\textwidth}
\includegraphics[width=\textwidth]{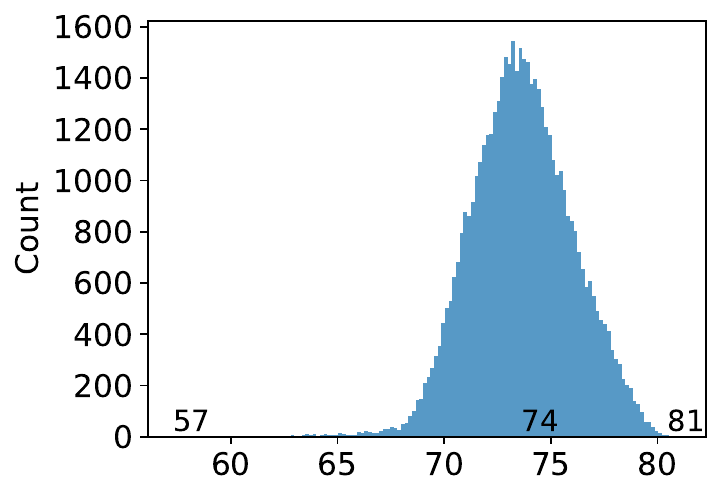}
\caption{Angle Degree ($\,^\circ$)}
\label{fig:angle_distribution}
\end{subfigure}
\caption{Distribution of tangential components and angle degrees. The adversarial examples were generated based on a VGG-16 model trained on clean CIFAR-10. The maximum, mean, and minimum values of the tangential component are 0.19, 0.34, and 0.58, respectively, while the corresponding values for the angle degrees are 57$^\circ$, 74$^\circ$, and 81$^\circ$.}
\label{fig:angle_tan_distribution}
\end{figure}

Building on this foundation, we present a new insight into how the geometry of adversarial perturbations contributes to the degradation of clean accuracy in adversarial training. Figure~\ref{fig:angle_tan_distribution} illustrates the distribution of tangential components and their corresponding angles within adversarial examples, where the angle of $\vx_i^*$ is defined between the tangent space at $\vx_i$ and the perturbation vector $\vx_i^*-\vx_i$. Our observations reveal that adversarial examples exhibit diverse tangential components and angles, suggesting that these geometric properties are critical factors in adversarial training. We formally argue that training on adversarial examples with a large normal component can significantly distort the decision boundary, ultimately leading to a substantial drop in clean accuracy. This insight, which has not been explored in prior work, serves as the conceptual basis for our proposed framework, \textit{\textbf{TA}ngent di\textbf{R}ection guided adversarial \textbf{T}raining} (TART). TART decomposes each adversarial perturbation into its tangent and normal components and adaptively adjusts the perturbation bound based on the tangential component. By avoiding training on off-manifold examples with large normal components, TART preserves a cleaner decision boundary and achieves improved clean accuracy on unperturbed clean data. An overview of our approach is illustrated in Figure~\ref{fig:TART}.


Our main contributions are summarized as follows:
\begin{itemize}
    \item We introduce \textit{Tangent Direction Guided Adversarial Training} (TART), the first adversarial training framework to explicitly leverage the tangent space and direction of adversarial examples.
    \item We provide empirical insights showing that adversarial examples with large normal components can significantly distort the decision boundary, which motivates the tangent-space-guided design of TART.
    \item We describe a practical method for estimating the tangent space of the data manifold and computing the tangential component of adversarial examples.
    \item We demonstrate through extensive experiments on simulated and benchmark datasets that TART consistently boosts clean accuracy without degrading robustness. Furthermore, we illustrate that TART is a universal framework that can be seamlessly integrated with a variety of existing adversarial defense methods, further enhancing their performance.
\end{itemize}

\section{Standard Adversarial Training}\label{sec:prelim}
In this section, we provide an overview of standard adversarial training~\citep{madry2018towards} and its implementation.

\subsection{Notation}
We focus on multiclass classification problems with $c$ classes. Let $\mathcal{D}$ be the data manifold and $\gX$ the input feature space. The training dataset $\{(\vx_i,y_i)\}^n_{i=1}$ is sampled from $\gD$, where $\vx_i \in \gX$ and $y_i \in \gY = \left\{ 1, \dots, c \right\}$. We assume $\gX = \R^d$ and define a metric space $(\gX, \left\| \cdot \right\|_p)$ based on the $\ell_p$ norm. Let $\gB_p(\epsilon)= \{ \vdelta \in \gX : \left\| \vdelta \right\|_p \leq \epsilon \}$ denote the closed $\epsilon$-ball centered at the origin. We consider a classifier $f(\cdot\,;\theta):\gX \rightarrow \R^c$ parametrized by $\theta$, where the $i$-th element of the output is the score of the $i$-th class. The goal  is to train a classifier $f$ that minimizes $\E_{(\vx,y)\sim \gD} [ \ell(f(\vx;\theta),y)]$, where $\ell:\R^c \times \gY \rightarrow \R$ is the classification loss.

\subsection{Learning Objective}
\citet{madry2018towards} proposed \textit{standard adversarial training} (standard AT), which solves a min-max optimization problem with the following objective function:
\begin{equation}
\label{eq:AT_loss}
    \min_\vtheta \frac{1}{n} \sum_{i=1}^n \ell(f(\vx^*_i;\vtheta),y_i),
\end{equation}
where
\begin{equation}
\label{eq:AT_generate_adversarial}
    \vx^*_i = \vx_i +  \argmax_{\vdelta \in \gS} \ell(f(\vx_i+\vdelta;\vtheta),y_i)
\end{equation}
and $\gS$ is the adversarial set, the set of allowed perturbations. $\vx^*_i$ is also called an adversarial example of $\vx$ that maximizes the loss. In the adversarial training literature, the adversarial set is typically considered to be an $l_p$ norm-bounded closed ball, i.e., $\gS = \gB_p(\epsilon)$. Our paper, much like other previous studies, focuses on the $l_\infty$ norm constraint, but we note that our method can be adapted for use with alternative norms as well. Standard AT improves the model's robustness by training on adversarial examples. Unfortunately, this often comes at the cost of a significant reduction in accuracy on clean data~\citep{madry2018towards}.

\subsection{Projected Gradient Descent}
The objective function of adversarial training involves two main steps: generating adversarial examples (inner maximization; Equation~(\ref{eq:AT_generate_adversarial})) and minimizing the loss on the generated examples with respect to $\vtheta$ (outer minimization; Equation~(\ref{eq:AT_loss})). The most widely used approach for the inner maximization is the \textit{projected gradient descent} (PGD) method, introduced by \citet{madry2018towards}. Standard AT also employs the PGD method to obtain an approximate solution for the inner maximization problem. Given a natural data point $\vx^{(0)}$, we find an adversarial example by iteratively computing the following:
\begin{equation}\label{eq:pgd}
    \vx^{(t+1)} \leftarrow \Pi_{\vx^{(t)} + \gS} \left( \vx^{(t)} + \alpha \sign \left(\nabla_{\vx^{(t)}} \ell (f(\vx^{(t)}; \vtheta), y ) \right) \right)
\end{equation}
where $\vx^{(t)}$ is the adversarial example at step $t$, $y$ the class label of $\vx^{(0)}$, $\Pi$ the projection operation, $\alpha$ the step size, and $\sign$ the sign function. We refer to this procedure with $k$ iterations as PGD$^k$ throughout the paper.
\section{Tangent Direction Guided Adversarial Training (TART) }
\label{sec:tangentat}

In this section, we propose \textit{TAngent diRection guided adversarial Training} (TART) and its implementation. TART is centered on the geometric intuition that perturbations with large normal components can over-distort the decision boundary, ultimately reducing a model's accuracy on clean data.

\begin{figure}[t]
\centering
\includegraphics[width=0.45\textwidth, trim={0.3cm 0.2cm 0.2cm 0.2cm},clip]
{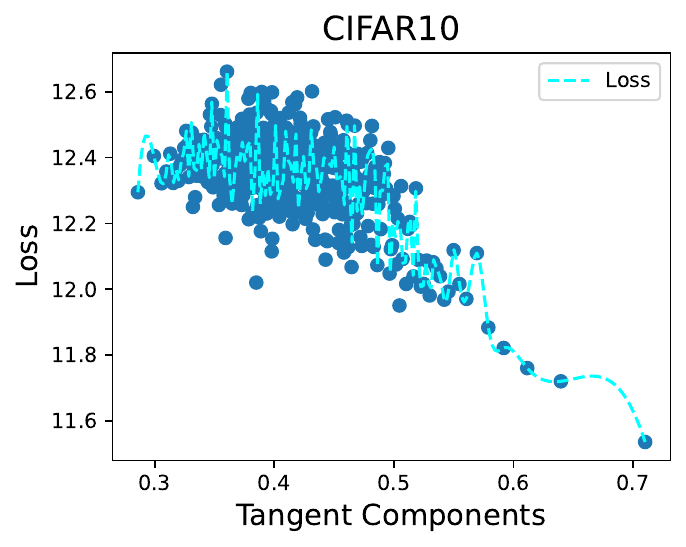}
\caption{Loss vs. Mean of tangential components within a batch.}
\label{fig:tart_loss}
\end{figure}

To verify this intuition, we first conduct an exploratory analysis investigating the relationship between adversarial perturbations and model loss. Figure~\ref{fig:tart_loss} illustrates the batch loss relative to the average tangential components of adversarial examples (generated using a VGG-16~\citep{Simonyan2014VeryDC} model on CIFAR-10~\citep{Krizhevsky2009LearningML}). We observe a clear trend: as the tangential component increases (i.e., the perturbation stays closer to the data manifold), the batch loss tends to decrease. This suggests that adversarial examples with larger tangential components are less harmful to the clean data distribution, whereas those with large normal components significantly increase the loss and distort the decision boundary. These findings provide the primary motivation for TART. Instead of using the uniform adversarial sets common in standard AT, TART employs a data-specific perturbation set that leverages information from the data manifold and tangent space to better preserve clean data characteristics.

To implement TART, we first craft an adversarial example $\vx_i^*$ of $\vx_i$ with the adversarial set $\gS= \gB_p(\epsilon)$ as described in Equation~(\ref{eq:AT_generate_adversarial}). Next, we estimate the tangent space at $\vx_i$ and then compute the \textit{tangential component} of $\vx_i^*$, which is defined as the norm of its projection onto the tangent space. TART assigns an adaptive perturbation limit $\epsilon_i$ to each adversarial example based on its tangential component. Adversarial examples with a larger tangential component receive a larger $\epsilon_i$, while those with a smaller tangential component receive a smaller $\epsilon_i$.
Using this adaptive bound, we generate the final adversarial data $\vx^{**}_i$ for training:
\begin{equation}\label{eq:TART_generate_adversarial}
    \vx^{**}_i =  \vx_i + \argmax_{\vdelta \in \gS_i} \ell(f(\vx_i+\vdelta;\vtheta),y_i),
\end{equation}
where $\gS_i = \gB_p(\epsilon_i)$. Equation~(\ref{eq:TART_generate_adversarial}) can be seen as a generalization of standard AT, as it recovers Equation~(\ref{eq:AT_generate_adversarial}) when $\epsilon_i = \epsilon$ for all $i$.

The overall procedure of TART is illustrated in Figure~\ref{fig:TART}. 
In the following subsections, we discuss in detail how we approximate the tangent space~(Section \ref{subsec:find_tangent_space}), compute the tangential component~(Section \ref{subsec:compute_tan_angle}), and choose an adaptive perturbation bound used for training~(Section \ref{subsec:sel_eps}).

\subsection{Finding the Tangent Space} \label{subsec:find_tangent_space}

\begin{figure}[t]
\centering
\begin{tikzpicture}[thick,scale=1.0, every node/.style={transform shape}]
    \begin{scope}[fill=red]
    \draw[ultra thick] (0,0) to [bend left=30] (3.5,4);
    \draw[ultra thick] (3.5,4) to [bend left=20] (6.2,1.5) node [label={[label distance=0.7cm, xshift=0.2cm]:$\mathcal{D} \in \mathbb{R}^d$}] {};
    \draw[ultra thick] (6.2,1.5) to [bend right=20] (2,-0.6);
    \draw[ultra thick] (2,-0.6) to [bend right=30] (0,0);
    \end{scope}
    
    \draw[very thick, smooth cycle, tension=0, color={rgb:red,5;green,5;blue,10}] plot coordinates{(2,1.5) (2.5, 2.7) (4,2.8) (3.6, 1.7)} node [label={[label distance=-0.9cm, xshift=-2cm]:tangent space}] {};
    \node[label={[label distance=-0.6cm, xshift=0.1cm]:$\vx$}] (B) at (3,2.1) {$\bullet$};
    \draw[very thick, dashed, color=purple] (2.5,1.4) to [bend left=30] (4.2,2.6) node [label={[label distance=-0cm, xshift=-.4cm]:$\{\mathbf{D}(\vz_i)\}$}] {} ;
    \path[very thick, color=purple, ->] (3,2.1) edge node[left, pos=0.7] {$\vv$} (3.6, 2.7);
    
    \draw[ultra thick, ->] (4, -2.5) -- (6.5, -2.5) ;
    \draw[ultra thick, ->] (4, -2.5) -- (4, -0.5) node [label=right:$\mathbb{R}^k$] {};
    \node[label={[label distance=-0.7cm, xshift=0.1cm]:$\vz$}] (A) at (5.2,-1.8) {$\bullet$};
    \draw[very thick, dashed, color=purple] (4.5,-1.8) to (5.9,-1.8) node [label={[label distance=-0.15cm, xshift=0.1cm]:$\{\vz_i\}$}] {} ;

    \path[->] (4, -0.3) edge [bend left] node[left, pos=0.25] {$\mathbf{D}$} (3, 1.3);
    \path[->] (3.5, 1.5) edge [bend left] node[right, pos=0.7] {$\mathbf{E}$} (4.5, 0);
\end{tikzpicture}
\caption{Illustration of Tangent Space Estimation. See  Algorithm~\ref{alg:compute_tan_angle} for a detailed description.}
\label{fig:tangent_illustration}
\end{figure}

For datasets where the data manifold is known, the tangent space can be obtained explicitly and used directly in TART. However, for benchmark or real-world datasets, the corresponding data manifold is often unknown, and we must therefore estimate the tangent space.

Suppose a dataset $\{\vx\} \subset \R^d$ lies on a $k$-dimensional manifold embedded in $\R^d$, where $k<d$. Indeed, many studies have demonstrated that the intrinsic dimension of image datasets is considerably smaller than their pixel space dimension~\citep{Levina2004MaximumLE, pope2021the}. Let $\mathbf{D}:\R^k \rightarrow \R^d$ be a parameterization of this $k$-dimensional manifold. For a data point $\vx \in \R^d$, let $\vz \in \R^k$ be its corresponding parameterization, such that $\mathbf{D}(\vz) = \vx$. By a first-order Taylor approximation, we can approximate the manifold locally as a hyperplane:
\begin{equation}
    \mathbf{D}(\vz + \vdelta) \approx \mathbf{D}(\vz) + \mathbf{J}_{\mathbf{D}}(\vz) \vdelta
\end{equation}
where $\vdelta \in \R^k$ and $\mathbf{J}_{\mathbf{D}}$ is the Jacobian matrix of $\mathbf{D}$. Therefore, a tangent vector at $\vx$ can be approximated by $\mathbf{D}(\vz + \delta \ve_i) - \mathbf{D}(\vz)$, where $\ve_i \in \R^k$ is a unit vector. To obtain a better approximation of the tangent vector, we can sample multiple points $\{\vz + \delta \ve_i \}$ around $\vz$ along each dimension $i$ with different $\delta$s. We then perform PCA on these sampled points and use the first principal component as an approximation of the $i$-th tangent vector. By repeating this procedure for each dimension $i = 1,...,k$, we obtain a set of $k$ tangent vectors that span the tangent space at $\vx$.

Unfortunately, the true intrinsic data manifold is unknown for most image datasets. Therefore, before approximating the tangent space, we must first learn a parameterization of the manifold, denoted as $\mathbf{D}$. A common approach is to use an autoencoder trained to reconstruct the input data~\citep{Bank2020Autoencoders}. Refer to Algorithm~\ref{alg:compute_tan_angle} for the steps estimating the tangent space, and see Figure~\ref{fig:tangent_illustration} for a visual representation.

\begin{algorithm}[t]
\caption{Finding the Tangent Space}
\textbf{Input:} \multiline{Natural data $\vx \in \R^d$, Adversarial data $\vx^*\in \R^d$,\\ Encoder $\mathbf{E}: \R^d \rightarrow \R^k $, Decoder $\mathbf{D}:\R^k \rightarrow \R^d$ }\\
\textbf{Output:} Tangent space at $\vx$
\begin{algorithmic}[1]
\State $\vz \leftarrow \mathbf{E}(\vx)$ 
\For{$i=1,\dots,k$}
\State Sample $\vz_1,\dots,\vz_l$ around $\vz$ along $i$-th dimension
\State Obtain $\mX'=(\vx'_1,\vx'_2,\dots,\vx'_l)$ where $\vx'_j = \mathbf{D}(\vz_j)$
\State \multiline{Do PCA on $\mX'$ and store the first principal component $\va_i \in \R^d$}
\EndFor
\State Obtain $\mA = (\va_1,\va_2,\dots,\va_k) \in \R^{d\times k}$ where the $k$-dimensional tangent space at $\vx$ is the column space of $\mA$ 
\end{algorithmic}
\label{alg:compute_tan_angle}
\end{algorithm}

\subsection{Computing Tangential Component}\label{subsec:compute_tan_angle}

Using the estimated tangent space, we compute the tangential component of an adversarial example. Let $\mA$ be a $d \times k$ matrix whose columns are the tangent vectors of a natural data $\vx\in\R^d$. We note that the column space of $\mA$ represents the $k$-dimensional tangent space at $\vx$. Given an adversarial example $\vx^*$ obtained by perturbing $\vx$, we define $\vw$ as the projection of the perturbation $\vx^* - \vx$ onto the tangent space. This projection can be computed using the projection matrix $\Pi_\mA = \mA(\mA^\intercal \mA)^{-1}\mA^\intercal$. A more detailed explanation of this projection is provided in Appendix~\ref{app:projection}. Therefore, the tangential component of $\vx^*$ is calculated as:
\begin{align}
\begin{split}
   \left\| \vw \right\| & = \left\|\Pi_\mA(\vx^* - \vx) \right\| \\
   & = \left\|\mA(\mA^\intercal \mA)^{-1}\mA^\intercal(\vx^*-\vx) \right\|     
\end{split}
\label{eq:tan}
\end{align}

\subsubsection{Saving Time Complexity of TART}

TART requires the estimation of the tangent space and the matrix calculation $\mA(\mA^\intercal \mA)^{-1}\mA^\intercal$ for each training data point in every epoch. To reduce this significant computational overhead, we pre-compute the tangent space and the corresponding matrix $\mA(\mA^\intercal \mA)^{-1}\mA^\intercal$ for each training image before training begins and store them. One limitation of pre-saving matrices is that it restricts the use of data augmentation techniques such as random cropping or horizontal flips, as such operations would randomly alter the data manifold. 

As a result of this strategy, we were able to minimize the computational burden during training. TART's training time only increased by a factor of 1.24 compared to standard AT, which is a very reasonable overhead in line with other advanced defense techniques~\citep{zhang2021GAIRAT, Zhang2019TRADES, Wang2020MART}. For details on training time measurements and the experimental setup, see Appendix~\ref{app:train_time}.

\subsubsection{Saving Space Complexity of TART}
Storing matrices to reduce training time requires a considerable amount of storage space. For example, the train set of CIFAR-10 consists of 50,000 RGB color images, each with a dimension $d=\;$3,072. Consequently, the matrix $\mA(\mA^\intercal \mA)^{-1}\mA^\intercal$ becomes a large $d\times d$ dense matrix. To address this, instead of saving a $d\times d$ matrix $\mA(\mA^\intercal \mA)^{-1}\mA^\intercal$, we suggest storing two $d\times k$ matrices $\mA(\mA^\intercal \mA)^{-1}$ and $\mA$. This provides significant space efficiency, as the intrinsic dimension $k$ is considerably smaller than the pixel space dimension $d \; (k \ll d)$.

\begin{algorithm}[t]
\caption{TART}

\textbf{Input:} \multiline{Training dataset $\{(\vx_i,y_i)\}^n_{i=1}$, Perturbation $\epsilon_{\max}$} 
\textbf{Output:} \multiline{Trained deep network $f(\vx ; \vtheta)$}
\begin{algorithmic}[1]
\State Train Autoencoder $(\mathbf{E}, \mathbf{D})$ on the training set
\State Compute tangent space of each training data using Algorithm~\ref{alg:compute_tan_angle} and store them for line 8
\For{epoch $=1,\dots,T$}
\For{mini-batch $=1,\dots,M$}
\State Sample a mini-batch $\mathcal{B}=\{(\vx_i,y_i)\}^m_{i=1}$ from $\mathcal{D}$
\For{$i=1,\dots,m$}
\State \multiline{Generate adversarial example $\vx^*_i$ of $\vx_i$ with perturbation bound of $\epsilon_{\max}$}
\State \multiline{Calculate tangential component of $\vx^*_i$ by Equation~(\ref{eq:tan}) using the stored tangent space information in line 2}
\EndFor
\State Assign $\epsilon_i$ according to Equation~(\ref{eq:eps_assign})
\State \multiline{$\vx^{**}_i = \vx^*_i$ if $\epsilon_i = \epsilon_{\max}$ \\ $\vx^{**}_i = \vx_i$ if $\epsilon_i = 0$ for all $1 \leq i \leq m$}
\State \multiline{$\vtheta \leftarrow \vtheta - \eta \nabla_\vtheta \left\{ \sum_{i=1}^m \frac{1}{m} l(f(\vx^{**}_i),y_i)\right\}$}
\EndFor
\EndFor
\end{algorithmic}
\label{alg:tart_training}
\end{algorithm}

\subsection{Selecting the Perturbation Bound $\epsilon$}\label{subsec:sel_eps}
The final step in realizing TART is to select an adaptive perturbation bound, $\epsilon_i$, for each data during training. We first generate adversarial examples with a perturbation bound of $\epsilon_{\max}$ for each training data. Next, the tangential components computed in Section~\ref{subsec:compute_tan_angle} are used as the criterion for choosing its final $\epsilon$. The fundamental idea is to assign larger or smaller $\epsilon$ values to data with larger or smaller tangential components. While several assignment methods exist, TART adopts a straightforward approach providing $\epsilon_{\max}$ to the upper 50\% and 0 to the lower 50\%. A brief overview of alternative assignment strategies and their results is provided in Appendix~\ref{app:epsilon_assignment}. The perturbation bound of the $i$-th data in a mini-batch can thus be expressed as:
\begin{equation}\label{eq:eps_assign}
\epsilon_i = \mathbb{I}\left( \,TC_i \geq \text{median}\{TC_k\}_{k=1}^B \, \right) \cdot \epsilon_{\max}
\end{equation}
where $TC_i$ is the tangential component of the $i$-th data and $B$ is the mini-batch size. The main advantage of this simple assignment is that it avoids the need to regenerate adversarial images. When $0 <\epsilon_i < \epsilon_{\max}$, a new adversarial example must be created with the new bound $\epsilon_i$. However, by only using bounds of 0 or $\epsilon_{\max}$, we can reuse existing images: the original image for a bound of 0, and the pre-computed adversarial image for a bound of $\epsilon_{\max}$. This significantly reduces the training time, as generating adversarial examples is computationally expensive. The overall training process of TART is summarized in Algorithm~\ref{alg:tart_training}.

\subsubsection{Theoretical Justification for $\epsilon$ Selection}
We provide a theoretical explanation for selecting the perturbation bound $\epsilon$. We show that training models on adversarial examples with large normal components, which are far from the data manifold, can decrease clean accuracy performance. For simplicity, we focus on a binary classification problem with a mean absolute loss function $L$. Let the feature space be $\mathcal{X} \subset \mathbb{R}^{d}$, where data points and labels lie in $\mathcal{X} \times \left\{ 0, 1 \right\}$. We assume the distributions of the clean training set and test set are identical, denoted as $\gP$. The distribution of adversarial examples, on which we train the model, is denoted as $\gQ$. Let the true labeling function $h: \gX \rightarrow \left\{ 0, 1 \right\}$. In the adversarial training setting, $\gP$ and $\gQ$ share the same true labeling function. The goal of the classification problem is to learn a function $f: \gX \rightarrow \left\{ 0, 1 \right\} $ that minimizes the expected risk $\gR_{\gP}(f) := \mathbb{E}_{\vx \sim \gP} [L(f(\vx),h(\vx))]$. Here, we develop bounds on the difference between $\gR_{\gP}(f)$ and $\gR_{\gQ}(f)$:

\begin{proposition}\label{prop:bound}
For any function $f$,
$$|\gR_{\gP}(f) - \gR_{\gQ}(f)| \leq 4  \, \text{TV}(\gP, \gQ),$$
where $\text{TV}(\gP, \gQ)$ denotes the total variance distance between $\gP$ and $\gQ$.
\end{proposition}

The proof is provided in Appendix~\ref{app:proof}. The total variance distance is a distance measure for probability distributions. Proposition~\ref{prop:bound} suggests that when the distributions $\gP$ and $\gQ$ are similar, the model may achieve better clean accuracy. This indicates that we should avoid training models on adversarial examples with large normal components.

\subsubsection{Connection between TART and Margin}\label{app:margin_connection}

\begin{figure}[ht]
\centering
\begin{subfigure}{0.55\textwidth}
\includegraphics[width=\textwidth]{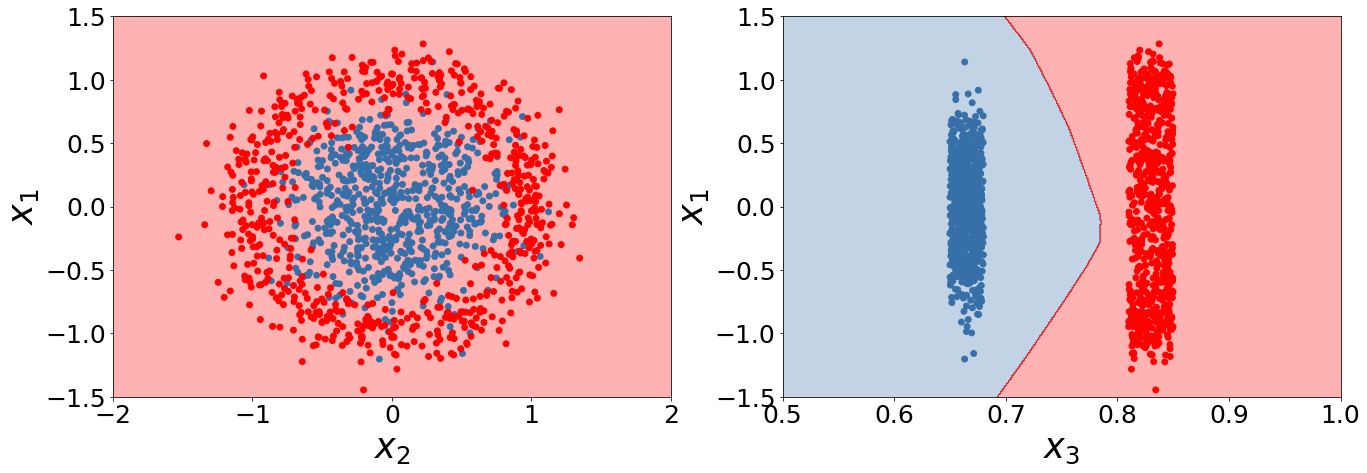}
\caption{Clean training}
\end{subfigure}
\begin{subfigure}{0.55\textwidth}
\includegraphics[width=\textwidth]{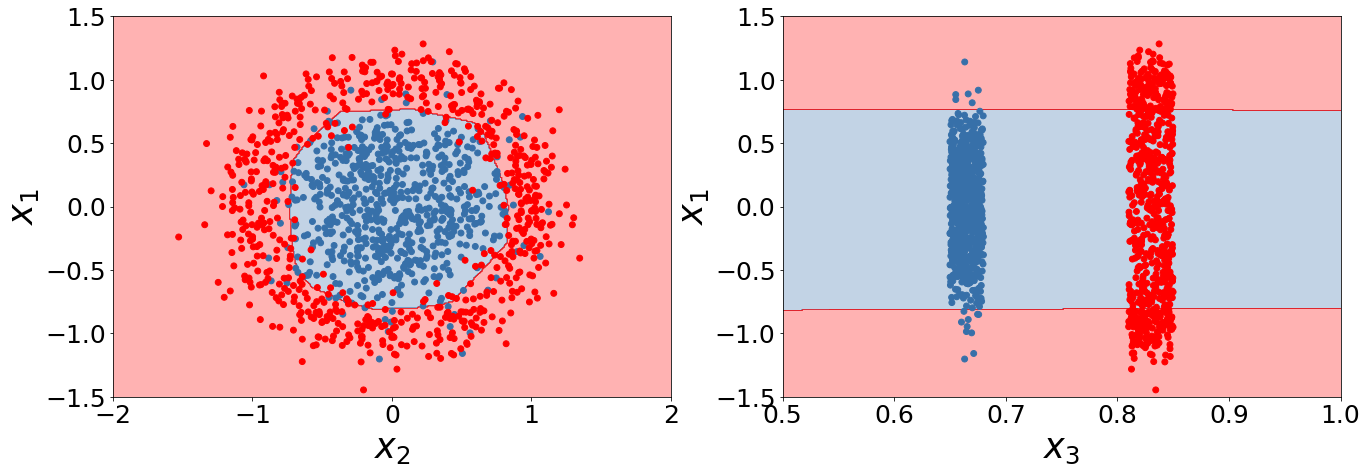}
\caption{Standard AT}
\end{subfigure}
\begin{subfigure}{0.55\textwidth}
\includegraphics[width=\textwidth]{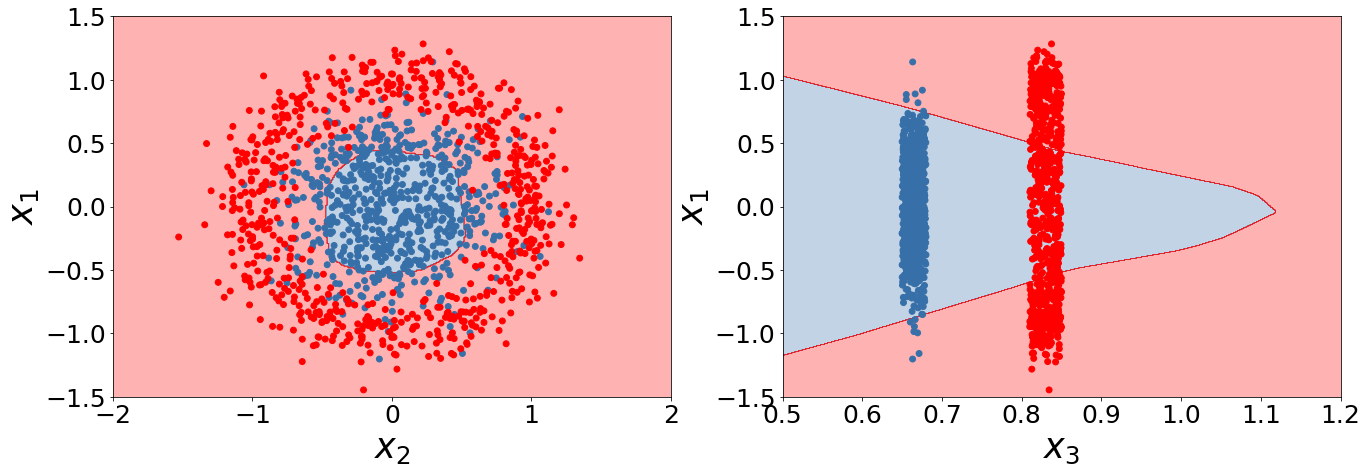}
\caption{TART}
\end{subfigure}
\caption{Decision boundary visualization for the toy problem by \citet{Rade2022HATReducingEM}: (Left) $x_3 = 0.85$, (Right) $x_2=0$. Standard AT considerably enhances robustness from 56\% to 77\% but results in a reduction in accuracy from 100\% to 92\%. TART recovers the accuracy to 96\% and even slightly improves the robustness to 79\%.}
\label{fig:tart_margin}
\end{figure}

\citet{Rade2022HATReducingEM} demonstrated that adversarial training results in an excessive increase in the margin along certain adversarial directions, which leads to a decrease in clean accuracy. While our approach does not directly optimize for the margin, its tangent space perspective provides an inherent mechanism for preventing this undesired margin increase. This is because TART refrains from training on large normal components. We illustrate this by utilizing a toy example proposed by \citet{Rade2022HATReducingEM}. This example is a 3D binary classification problem where two classes lie in two noisy concentric circles parallel to the $x_1x_2$-plane but are separable along the $x_3$ dimension. 

As shown in Figure~\ref{fig:tart_margin}, we use clean training, standard AT, and TART to train a single hidden layer multilayer perceptron and compare the resulting decision boundaries. Standard AT achieves robustness at the expense of an unbounded margin in the $x_3$ direction, resulting in a loss of clean accuracy. In contrast, TART maintains a finite margin and improves clean accuracy without compromising robustness. 
These results not only validate our insight but also highlight TART’s potential to mitigate undesirable geometric side effects of adversarial training. By implicitly regularizing the decision boundary through a tangent-aware perturbation scheme, TART preserves both robustness and generalization, demonstrating a promising direction for geometry-guided defenses.

\begin{table*}[t]
\begin{center}
{
\footnotesize
\begin{tabular}{c|c|cc|cc|cc} \toprule
\multirow{2}{*}{Method} & \multirow{2}{*}{Accuracy} & \multicolumn{2}{c|}{$\R^{100} \; (c=4)$} & \multicolumn{2}{c|}{$\R^{100} \; (c=8)$}  & \multicolumn{2}{c}{$\R^{100} \; (c=16)$} \\ 
&  & $\epsilon=\;$0.03 & $\epsilon=\;$0.05 & $\epsilon=\;$0.03  & $\epsilon=\;$0.05&  $\epsilon=\;$0.01 & $\epsilon=\;$0.03\\  \midrule \midrule
\multirow{2}{*}{TART} & Clean  & \textbf{95.3 $\pm$ 0.8} &  \textbf{93.5 $\pm$ 1.5} & \textbf{93.2 $\pm$ 2.1} & \textbf{88.2 $\pm$ 2.5} & \textbf{94.0 $\pm$ 0.8} & \textbf{85.6 $\pm$ 1.7} \\
 & Robust &  \textbf{63.1 $\pm$ 2.6} &  \textbf{48.1 $\pm$ 3.8} & \textbf{40.6 $\pm$ 1.4} & \textbf{24.8 $\pm$ 1.6} &  \textbf{49.3 $\pm$ 2.4} & \textbf{18.0 $\pm$ 1.8} \\
 \midrule
\multirow{2}{*}{Reverse-TART} & Clean  & 91.9 $\pm$ 3.7 &  87.9 $\pm$ 4.8 & 87.9 $\pm$ 1.7 & 82.5 $\pm$ 1.3 & 89.4 $\pm$ 1.3 & 82.2 $\pm$ 3.4 \\
 & Robust &  57.5 $\pm$ 2.1 &  43.9 $\pm$ 3.7 & 39.8 $\pm$ 1.1 &  23.5 $\pm$ 2.3 & 46.9 $\pm$ 2.5 & 15.4 $\pm$ 1.9 \\ \toprule
 \multirow{2}{*}{Method} & \multirow{2}{*}{Accuracy} & \multicolumn{2}{c|}{$\R^{200} \; (c=4)$} & \multicolumn{2}{c|}{$\R^{200} \; (c=8)$}  & \multicolumn{2}{c}{$\R^{200} \; (c=16)$} \\ 
&  & $\epsilon=\;$0.03 & $\epsilon=\;$0.05 & $\epsilon=\;$0.03  & $\epsilon=\;$0.05&  $\epsilon=\;$0.01 & $\epsilon=\;$0.03\\  \midrule \midrule
\multirow{2}{*}{TART} & Clean  & \textbf{95.9 $\pm$ 1.4} &  \textbf{90.7 $\pm$ 1.0} & \textbf{91.4 $\pm$ 1.6} & \textbf{88.0 $\pm$ 1.0} & \textbf{93.2 $\pm$ 1.6} & \textbf{85.4 $\pm$ 1.8} \\
 & Robust &  \textbf{54.7 $\pm$ 2.2} &  \textbf{37.9 $\pm$ 4.1} & \textbf{29.5 $\pm$ 3.8} & \textbf{16.5 $\pm$ 1.6} & \textbf{39.3 $\pm$ 0.5} & \textbf{9.4 $\pm$ 0.9} \\
 \midrule
\multirow{2}{*}{Reverse-TART} & Clean  & 90.1 $\pm$ 3.0 &  87.2 $\pm$ 2.0 & 88.0 $\pm$ 1.7 & 78.1 $\pm$ 2.6 & 88.9 $\pm$ 1.1 & 77.8 $\pm$ 2.8 \\
 & Robust &  50.4 $\pm$ 3.1 &  37.6 $\pm$ 3.2 & 28.3 $\pm$ 2.9 &  12.7 $\pm$ 3.6 & 38.8 $\pm$ 1.8 & 7.9 $\pm$ 0.7 \\ \toprule
  \multirow{2}{*}{Method} & \multirow{2}{*}{Accuracy} & \multicolumn{2}{c|}{$\R^{400} \; (c=4)$} & \multicolumn{2}{c|}{$\R^{400} \; (c=8)$}  & \multicolumn{2}{c}{$\R^{400} \; (c=16)$} \\ 
&  & $\epsilon=\;$0.01 & $\epsilon=\;$0.03 & $\epsilon=\;$0.01  & $\epsilon=\;$0.03 &  $\epsilon=\;$0.01 & $\epsilon=\;$0.03\\  \midrule \midrule
\multirow{2}{*}{TART} & Clean  & \textbf{97.8 $\pm$ 0.3} &  \textbf{94.1 $\pm$ 1.4} & \textbf{96.9 $\pm$ 0.5} & \textbf{86.9 $\pm$ 0.8} & \textbf{91.4 $\pm$ 1.2} & \textbf{82.1 $\pm$ 4.0} \\
 & Robust &  \textbf{74.5 $\pm$ 0.6} &  \textbf{46.5 $\pm$ 1.8} & \textbf{54.1 $\pm$ 0.5} & \textbf{18.3 $\pm$ 1.5} & \textbf{26.3 $\pm$ 0.6} & \textbf{4.4 $\pm$ 0.9} \\
 \midrule
\multirow{2}{*}{Reverse-TART} & Clean  & 97.6 $\pm$ 0.7 &  84.5 $\pm$ 3.8 & 91.7 $\pm$ 0.7 & 82.9 $\pm$ 1.3 & 84.9 $\pm$ 1.7 & 70.5 $\pm$ 3.8 \\
 & Robust &  60.6 $\pm$ 1.6 &  41.2 $\pm$ 1.5 & 50.9 $\pm$ 1.4 &  17.3 $\pm$ 2.9 & 24.9 $\pm$ 1.6 & 4.0 $\pm$ 1.1 \\ 
\bottomrule
\end{tabular}}
\end{center}
\caption{Test accuracies (\%) on transformed hemisphere dataset under different dimensions $d$, number of classes $c$, and perturbation bounds $\epsilon$. The average clean and robust accuracies over five trials are reported along with their standard deviation. }\label{tab:sphere_exp}
\end{table*}

\section{Experiments}
\label{sec:experiments}

In this section, we present a series of experiments to validate the effectiveness of our proposed method, TART. We begin with a simulated experiment on a transformed hemisphere dataset, which serves to illustrate and validate our core theoretical insights. We then evaluate TART on the CIFAR-10 dataset, demonstrating that it not only improves clean accuracy but can also be effectively combined with existing adversarial training methods. Finally, we evaluate TART on the Tiny ImageNet dataset to assess its ability to generalize to more challenging and large-scale settings.

\subsection{Transformed Hemisphere}\label{subsec:hemisphere_experiment}

To compare the efficacy of training on adversarial examples with large normal components versus those with large tangential components, we conducted an experiment using a simulated dataset. Our dataset is based on a unit hemisphere in $\mathbb{R}^3$, where the tangent space can be computed without any approximation. The hemisphere is first evenly divided into $c$ regions to serve as class labels.

Data points are sampled from the hemispherical surface. Let $\vz \in \mathbb{R}^3$ be a sampled point, which is then transformed to a high-dimensional space $\mathbb{R}^d$ via a linear transformation, $\vx = T(\vz)$. Here, $T:\mathbb{R}^3 \to \mathbb{R}^d$ is a linear map whose columns are three orthonormal vectors in $\mathbb{R}^d$. The class label of $\vx$ is determined by the region where $\vz$ is located.
Since the data points lie on a spherical manifold, we can explicitly compute the tangent vectors, $\vu_1$ and $\vu_2$, of $\vz$. The corresponding tangent vectors for $\vx$ are then obtained by applying the transformation $T$ to $\vu_1$ and $\vu_2$. Using these tangent vectors, we can accurately calculate the tangential component of adversarial examples and implement TART on the simulated dataset.


The goal of this experiment is to confirm the validity of our core concepts and assess the effectiveness of TART. To this end, we compare TART to a contrasting approach, which we call Reverse-TART. Reverse-TART provides a smaller perturbation bound to adversarial examples with larger tangential components, directly opposing the core hypothesis of our method. To obtain a more pronounced effect, we compute the tangential components at each epoch and use only a subset of the adversarial examples: those with the largest 25\% and the smallest 25\% tangential components. The perturbation bound for each adversarial example during training is determined as follows:
\begin{itemize}
    \item TART: provide $\epsilon$ for the largest 25\% and 0 for the smallest 25\% as their perturbation bounds.
    \item Reverse-TART: provide 0 for the largest 25\% and $\epsilon$ for the smallest 25\% as their perturbation bounds.
\end{itemize}

We train a two hidden layer DNN across various experimental settings: dimension $d \in \{ 100, 200, 400\}$, number of classes $c \in \{ 4, 8, 16\}$, maximum perturbation $\epsilon \in \{ 0.01, 0.03, 0.05\}$. The models are trained for 50 epochs using stochastic gradient descent (SGD) with a momentum of 0.9 and a weight decay of 0.0002. We use an initial learning rate of 0.1, which is decayed by a factor of 10 at epochs 30 and 45. The adversarial examples for training are generated by PGD$^{10}$ with a random start and a step size of $\alpha = \epsilon/4$. Note that a perturbation bound of 0 implies the use of natural images without any PGD attacks.

In Table~\ref{tab:sphere_exp}, we compare the performance of TART and Reverse-TART based on clean and robust test accuracy. Clean accuracy is the model's performance on the test dataset without attacks, while robust accuracy is measured against adversarial data generated by PGD$^{20}$ with a random start, an $l_\infty$ perturbation bound of $\epsilon$, and a step size of $\alpha = \epsilon/10$. Our results show that TART consistently outperforms Reverse-TART in both clean and robust accuracy across all cases. These findings support our core hypothesis that avoiding training on adversarial examples with large normal components is an effective strategy for enhancing clean accuracy without sacrificing robustness.

\begin{table*}[t]
\caption{Test accuracies (\%) on CIFAR-10. The performances over three trials are reported along with their standard deviation.}\label{tab:cifar10_experiment}
\begin{center}
{\scriptsize 
\begin{tabular}{c|cc|cccc} \toprule
Defense & Clean (Last) & Clean (Best) & FGSM & PGD$^{20}$  & PGD$^{40}$ & AutoAttack\\  \midrule \midrule
Clean & 90.72 $\pm$ 0.10  &  90.86 $\pm$ 0.10 &  5.19 $\pm$ 1.03 &  0.0 $\pm$ 0.0  &  0.0 $\pm$ 0.0 &  0.0 $\pm$ 0.0 \\
\midrule
AT &  81.53 $\pm$ 0.21  &  81.76 $\pm$ 0.22 &  54.11 $\pm$ 0.37 &  37.06 $\pm$ 0.27  &  36.73 $\pm$ 0.20 &  35.89 $\pm$ 0.19 \\
TART-AT & \textbf{83.47 $\pm$ 0.22}  &  \textbf{83.73 $\pm$ 0.15} &  \textbf{55.27 $\pm$ 0.25} &  \textbf{37.72 $\pm$ 0.03}  &  \textbf{37.41 $\pm$ 0.13} &  \textbf{36.58 $\pm$ 0.08} \\
 \midrule
TRADES &  79.47 $\pm$ 0.09  &  79.85 $\pm$ 0.18 &  54.53 $\pm$ 0.10 &  
39.08 $\pm$ 0.06  &  38.68 $\pm$ 0.06 &  37.80 $\pm$ 0.07 \\
TART-TRADES &  \textbf{81.60 $\pm$ 0.06}  &  \textbf{81.95 $\pm$ 0.08} &  \textbf{55.49 $\pm$ 0.47} &  \textbf{39.30 $\pm$ 0.26} &  \textbf{38.99 $\pm$ 0.27} &  \textbf{38.26 $\pm$ 0.30} \\
 \midrule
MART &  79.82 $\pm$ 0.39  &  80.18 $\pm$ 0.22 &  53.15 $\pm$ 0.17 &  36.76 $\pm$ 0.39  &  36.16 $\pm$ 0.44 &  34.28 $\pm$ 0.56 \\
TART-MART & \textbf{82.33 $\pm$ 0.28}  &  \textbf{84.31 $\pm$ 0.15} & \textbf{55.19 $\pm$ 0.12} &  \textbf{37.64 $\pm$ 0.25}  &  \textbf{37.17 $\pm$ 0.25} &  \textbf{35.86 $\pm$ 0.16} \\
 \midrule
GAIRAT &  81.37 $\pm$ 0.15  &  81.67 $\pm$ 0.12 &  54.13 $\pm$ 0.13 &  37.29 $\pm$ 0.23  &  36.89 $\pm$ 0.25 &  36.03 $\pm$ 0.24 \\
TART-GAIRAT & \textbf{83.14 $\pm$ 0.15}  &  \textbf{83.47 $\pm$ 0.20} &  \textbf{55.24 $\pm$ 0.38} &  \textbf{37.43 $\pm$ 0.26}  &  \textbf{37.04 $\pm$ 0.32} &  \textbf{36.26 $\pm$ 0.36} \\
\bottomrule
\end{tabular}}
\end{center}
\end{table*}

\begin{table*}[t]
\caption{Test accuracies (\%) on Tiny ImageNet. The average performances over three trials are reported.}\label{tab:tiny_imagenet_experiment}
\begin{center}
{ \scriptsize 
\begin{tabular}{c|cc|cccc} \toprule
Defense & Clean (Last) & Clean (Best) & FGSM & PGD$^{20}$  & PGD$^{40}$ & AutoAttack\\  \midrule \midrule
Clean & 55.97 $\pm$ 0.21  & 58.01 $\pm$ 0.11 &  13.58 $\pm$ 0.35 &  7.50 $\pm$ 0.29  &  6.05 $\pm$ 0.40 &  0.32 $\pm$ 0.02 \\
\midrule
AT &  49.94 $\pm$ 0.19  &  53.18 $\pm$ 0.53 &  33.41 $\pm$ 0.21 &  26.50 $\pm$ 0.16 &  25.83 $\pm$ 0.16 &  \textbf{25.34 $\pm$ 0.18} \\
TART-AT & \textbf{53.19 $\pm$ 0.07}  &  \textbf{56.04 $\pm$ 0.20} &  \textbf{33.74 $\pm$ 0.09} &  \textbf{26.56 $\pm$ 0.26}  &  \textbf{25.87 $\pm$ 0.28} &  25.11 $\pm$ 0.08 \\
\bottomrule
\end{tabular}}
\end{center}
\end{table*}

\subsection{Performance Evaluation on CIFAR}\label{subsec:cifar}
In this section, we assess the performance of TART on CIFAR-10~\citep{Krizhevsky2009LearningML}. We also verify the versatility of TART by successfully integrating it with other existing defense methods. Specifically, we combine TART with the following defense approaches: (1) Standard AT~\citep{madry2018towards}, (2) TRADES~\citep{Zhang2019TRADES}, (3) MART~\citep{Wang2020MART}, and (4) GAIRAT~\citep{zhang2021GAIRAT}.

\subsubsection{Defense Settings}
We use WideResNet-32-10~\citep{Zagoruyko2016WideRN} (WRN-32-10) as in GAIRAT~\citep{zhang2021GAIRAT}, FAT~\citep{Zhang2020FAT}, and MAIL~\citep{Wang2021MAIL}. We train each defense model for 100 epochs using SGD with momentum 0.9, weight decay 0.0002. We initially use a learning rate of 0.1 and divide it by 10 at the 60th and 90th epochs. For the training attack, we use the PGD$^7$ attack with a random start, $l_\infty$ perturbation bound $\epsilon_{\max}=8/255$, and step size $\alpha = 2/255$. The hyperparameters featured in each method are configured to match those used in the original paper: $\beta = 6$ for TRADES, $\lambda = 5$ for MART. Also, only for MART, we followed using a maximum learning rate of 0.01 and weight decay of 0.0035 as in \cite{Wang2020MART}. Note that data augmentation techniques were not used during the experiment. All experiments were run on a single GeForce GTX 1080 GPU.

\subsubsection{Evaluation Metrics}
We report both the clean test accuracy at the best checkpoint and the last checkpoint. We also evaluate the robustness of all trained models against four types of adversarial attacks: Fast Gradient Sign Method (FGSM)~\citep{Goodfellow2015ExplainingAH}, PGD$^{20}$, PGD$^{40}$~\citep{madry2018towards}, and AutoAttack~\citep{croce2020reliable}. All attacks are constrained to $l_\infty$ perturbations with a maximum bound of $\epsilon_{\max} = 8/255$, i.e., $\left\| \vx - \vx^* \right\|_\infty \leq 8/255.$ The step size for PGD$^{20}$ and PGD$^{40}$ is set to $ \alpha = 2/255$, while FGSM is equivalent to a one-step PGD with step size $\epsilon$. FGSM, PGD$^{20}$, and PGD$^{40}$ attacks are conducted under a white-box setting where the attacker has full access to model parameters. AutoAttack is a combination of four different attack methods that include a black-box attack.

One major difference from the transformed hemisphere experiment discussed in Section~\ref{subsec:hemisphere_experiment} is the lack of knowledge regarding the exact tangent space of the data manifold. As described in Section~\ref{sec:tangentat}, we first train a convolutional autoencoder on CIFAR-10 to approximate the tangent space and compute the tangential components. See Appendix~\ref{app:autoencoder} for more details on the autoencoder. We leverage this knowledge to evaluate the performance of TART, as well as TART combined with well-known defense methods. Our experimental results are summarized in Table~\ref{tab:cifar10_experiment}. The results indicate that when TART is combined with existing methods, clean accuracy is consistently boosted, while robustness is maintained with no degradation. This again confirms the effectiveness of TART and suggests that examining the tangential component is a valuable approach for improving adversarial training.

For a more detailed analysis, we also performed ablation studies on the perturbation bound assignment strategies and the autoencoder latent dimension. The results of these experiments are provided in Appendices~\ref{app:autoencoder} and \ref{app:epsilon_assignment}.


\subsection{Performance Evaluation on Tiny ImageNet}
To further validate the effectiveness of TART on more complex datasets, we conducted experiments on Tiny ImageNet~\citep{le2015tiny}. The full experimental setup, including the model architecture and training hyperparameters, is detailed in Appendix~\ref{app:tiny_imagenet}.

As shown in Table~\ref{tab:tiny_imagenet_experiment}, our results on Tiny ImageNet are consistent with those observed on CIFAR-10. Compared to standard AT, TART achieves higher clean accuracy while maintaining comparable robustness. These findings suggest that TART’s ability to improve clean accuracy extends effectively to more complex and large-scale image classification tasks.
\section{Conclusion}\label{sec:conclusion}

In this paper, we introduced Tangent Direction Guided Adversarial Training (TART), a novel framework that leverages the tangent space of the data manifold to guide adversarial training. Our analysis revealed that adversarial examples with large normal components can excessively distort the decision boundary, thereby impairing generalization to natural data. TART mitigates this issue by adaptively modulating the perturbation bound based on the tangential component of each adversarial example. In particular, it assigns larger or smaller perturbation limits to adversarial examples with larger or smaller tangential components, respectively. 

Through experiments on both synthetic and benchmark datasets, we demonstrated that TART effectively improves clean accuracy while maintaining robustness comparable to standard adversarial training. To the best of our knowledge, TART is the first framework to incorporate tangent directions into adversarial learning. In future work, we plan to explore sample-wise optimal perturbation bounds by theoretically analyzing each example’s contribution to overall model performance. We believe that this tangent-space perspective opens new directions for geometry-aware adversarial defenses and provides a promising foundation for future research.









\bibliography{bibliography.bib}

\newpage
\appendix



\clearpage

\begin{center}
    \begin{spacing}{1.2}
        \LARGE \textbf{Supplementary Material: Improving Clean Accuracy via a Tangent-Space Perspective on Adversarial Training}
    \end{spacing}
    \large Bongsoo Yi, Rongjie Lai, Yao Li
\end{center}

\section{Related Work}\label{app:related_work}

This section provides an overview of recent advances in adversarial training that address the trade-off between clean accuracy and robustness. These methods are often motivated by the observation that deep neural networks, even when over-parameterized, suffer from limited effective capacity during adversarial training. \citet{zhang2021GAIRAT} empirically demonstrated that such models experience an excessive smoothing effect, which hinders their ability to fit both clean and adversarial data. To address this, various approaches have been proposed to make more efficient use of this limited capacity by treating individual training samples differently. These methods can broadly be categorized into two types: (1) reweighting the loss based on data difficulty, and (2) assigning adaptive perturbation bounds tailored to each sample.

One line of work~\citep{zhang2021GAIRAT, Wang2021MAIL, Wang2020MART, Kim2021EWAT} focuses on reweighting the loss function by assigning larger weights to important data points. \citet{zhang2021GAIRAT}~(GAIRAT) and \citet{Wang2021MAIL}~(MAIL) observed that data points with smaller margins, i.e., the distance from a data point to the model's decision boundary, are more prone to adversarial attacks. GAIRAT estimated the margin by counting the number of iterations in the projected descent method required to generate a misclassified adversarial example, while MAIL introduced the probabilistic margin which leverages the posterior probabilities of classes. Both methods then reweighted the loss function by giving greater importance to data points with smaller margins. \citet{Wang2020MART}~(MART) suggested that misclassified examples are crucial for enhancing robustness and therefore introduced additional regularization on such examples. \citet{Kim2021EWAT} (EWAT) reweighted the loss according to the entropy of the predicted distribution. 

Another line of work~\citep{Ding2020MMA, Cheng2020CATCA, Balaji2019IAATInstanceAA} that treats data unequally provides an adaptive perturbation bound instead of a fixed one. \citet{Ding2020MMA} (MMA) identified the margin for each data at each training epoch and utilized it as its perturbation magnitude. This allowed MMA to achieve margin maximization and improve the model's adversarial robustness. \citet{Cheng2020CATCA} (CAT) determined a non-uniform effective perturbation length by utilizing the customized training labels. Although not explicitly mentioned, \citet{Zhang2020FAT} (FAT) also employed adaptive perturbation bounds through training on friendly adversarial data that minimizes classification loss among the certainly misclassified adversarial examples. Driven by the motivation that using a common perturbation level may be inefficient and suboptimal, we also adopt varying perturbation bounds in this paper. However, in contrast to most margin-based studies, our approach does not directly rely on the margin. Instead, its tangent space perspective naturally prevents the undesired margin increase. See Section~\ref{app:margin_connection} for more details.

\section{Projection Matrix}\label{app:projection}
For a matrix $\mA$ of size $n\times k$ with full column rank, we define $\mathcal{C}_\mA$ as the column space of $\mA$, which is a linear subspace in $\mathbb{R}^n$ spanned by the columns of $\mA$. The column space $\mathcal{C}_\mA$ can be represented as:
\begin{align*}
\mathcal{C}_\mA & =\left \{ \beta_1\vu_1+\beta_2\vu_2+\cdots \beta_k\vu_k:\beta_1,\beta_2,\cdots,\beta_k \in \mathbb{R} \right \} \\
&=\left \{ \mA{\vbeta}:\vbeta\in \mathbb{R}^k \right \}  
\end{align*}
where $\vu_1, \vu_2, \cdots, \vu_k$ are the columns of $\mA$.

The projection matrix $\Pi_\mA := \mA(\mA^\intercal \mA)^{-1} \mA^\intercal$ is an $n \times n$ matrix that defines the linear operator projecting any vector $\vv \in \mathbb{R}^n$ onto the subspace $\mathcal{C}_\mA$. To derive this matrix, we use the fact that the residual vector $\vv - \Pi_\mA \vv$ must be orthogonal to $\mathcal{C}_\mA$. In addition, there exists a vector $\vbeta \in \mathbb{R}^k$ such that $\Pi_\mA\vv = \mA\vbeta$ since $\Pi_\mA\vv$ lies in the column space of $\mA$. Therefore, 
\begin{align*}
 & \quad\; \vv-\Pi_\mA\vv=\vv-\mA\vbeta\perp\mathcal{C}_\mA \\
&\Leftrightarrow \mA^\intercal(\vv-\mA\vbeta)=0 \\
&\Leftrightarrow \vbeta=(\mA^\intercal\mA)^{-1}\mA^\intercal\vv\\
&\Leftrightarrow\Pi_\mA=\mA(\mA^\intercal\mA)^{-1}\mA^\intercal.
\end{align*}
It is important to highlight that $\mA^\intercal\mA$ is always invertible when $\mA$ has full column rank, ensuring that the projection matrix $\Pi_\mA$ is well-defined.

\section{Proof of Proposition~\ref{prop:bound}}\label{app:proof}

\noindent\textbf{Proposition III.1.}
\textit{For any function $f$,
$$|\gR_{\gP}(f) - \gR_{\gQ}(f)| \leq 4  \, \text{TV}(\gP, \gQ),$$
where $\text{TV}(\gP, \gQ)$ denotes the total variance distance between $\gP$ and $\gQ$.}

\begin{proof}
Let $\gP \sim p$ and $\gQ \sim q$. Then, 
\begin{align*}
     & |\gR_{\gQ}(f) - \gR_{\gP}(f)| \\
     = & \left| \int L(f(\vx),h(\vx)) p(\vx) d \vx - \int L(f(\vx),h(\vx)) q(\vx) d \vx \right| \\
     \leq & \int  L(f(\vx),h(\vx)) \, |p(\vx) - q(\vx)| \, d\vx \\
     \leq & 2 \left\| \gP - \gQ \right\|_1 = 4  \, \text{TV}(\gP, \gQ),
\end{align*}
where the last equality holds due to the relationship between the $L^1$ norm and the total variance distance.
\end{proof}

\section{Training Time}\label{app:train_time}

We compare the average training time per epoch for five defense methods: standard AT, TART, GAIRAT~\citep{zhang2021GAIRAT}, TRADES~\citep{Zhang2019TRADES}, and MART~\citep{Wang2020MART}. All models are trained on CIFAR-10 using WideResNet-32-10 for 100 epochs with a single NVIDIA GeForce RTX 3090 GPU. As shown in Table~\ref{tab:TART_time}, TART takes approximately 1.24 times longer per epoch than standard AT. This marginal increase in training time is mainly due to the additional computation required for estimating the tangent space. The training time of TART is similar to, or slightly shorter than, that of other advanced defense methods such as GAIRAT, TRADES, and MART.

\begin{table}[h]
\centering
\begin{tabular}{c|c|c|c|c} \toprule
 AT & TART & GAIRAT & TRADES & MART \\  \midrule 
 406.12s & 506.66s &552.36s & 722.02s & 497.46s \\
\bottomrule
\end{tabular}
\caption{Training Time Comparison.}
\label{tab:TART_time}
\end{table}

\section{Experimental Setup on Tiny ImageNet}\label{app:tiny_imagenet}
Tiny ImageNet~\citep{le2015tiny} is a widely used benchmark in computer vision research. It contains a total of 120,000 images across 200 classes, with 500 training, 50 validation, and 50 test images per class. All images are 64 $\times$ 64 pixels, making it a more challenging and large-scale dataset than CIFAR-10.

\subsection{Defense Settings}
We adopt the VGG-16 architecture~\citep{Simonyan2014VeryDC} for all models. Training is performed for 100 epochs using SGD with momentum 0.9 and a weight decay of 0.0002. The initial learning rate is set to 0.1 and is reduced by a factor of 10 at the 60th and 90th epochs. For adversarial training, we apply a 7-step PGD (PGD$^7$) attack with a random initialization, an $l_\infty$ perturbation bound of $\epsilon_{\max} = 0.01$, and a step size of $\alpha = 0.002$. No data augmentation is used throughout the training. All experiments are conducted on a single NVIDIA GeForce GTX 1080 GPU.

\subsection{Evaluation Metrics} 
We evaluate the final model's performance based on clean test accuracy and robustness against three adversarial attacks: Fast Gradient Sign Method (FGSM)~\citep{Goodfellow2015ExplainingAH}, PGD$^{20}$, PGD$^{40}$~\citep{madry2018towards}, and AutoAttack~\citep{croce2020reliable}. All attacks operate within an $l_\infty$ perturbation bound of $\epsilon_{\max} = 0.01$. PGD$^{20}$ and PGD$^{40}$ use a step size of $\alpha = 0.002$, and FGSM is implemented as a single-step PGD with step size equal to $\epsilon$. Both FGSM and PGD are white-box attacks, assuming full access to the model. AutoAttack combines four diverse attacks to provide a comprehensive robustness assessment.

\section{Autoencoder Details}\label{app:autoencoder}
\vspace{2mm}

In this section, we describe the autoencoder architecture utilized for all experiments conducted on CIFAR-10 in Section~\ref{subsec:cifar}.
\begin{itemize}
    \item Encoder: 
    \begin{enumerate}
        \item Convolution layer with $4 \times 4$ filter, \# of input channels: 3, \# of output channels: 12, stride $= 2$, padding $= 1$
        \item ReLU activation layer
        \item Convolution layer with $4 \times 4$ filter, \# of input channels: 12, \# of output channels: 24, stride $= 2$, padding $= 1$
        \item ReLU activation layer
        \item Convolution layer with $4 \times 4$ filter, \# of input channels: 24, \# of output channels: 48, stride $= 2$, padding $= 1$
        \item ReLU activation layer
        \item Convolution layer with $4 \times 4$ filter, \# of input channels: 48, \# of output channels: 96, stride $= 2$, padding $= 1$
        \item ReLU activation layer
        \item Linear layer, \# of input channels: 384, \# of output channels: 128
    \end{enumerate}
    \item Latent: 128-dimensional
    \item Decoder:
    \begin{enumerate}
        \item Linear layer, \# of input channels: 128, \# of output channels: 384
        \item Transposed-Convolution layers with $4 \times 4$ filter, \# of input channels: 96, \# of output channels: 48, stride $= 2$, \\ padding $= 1$
        \item ReLU activation layer
        \item Transposed-Convolution layers with $4 \times 4$ filter, \# of input channels: 48, \# of output channels: 24, stride $= 2$, \\ padding $= 1$
        \item ReLU activation layer
        \item Transposed-Convolution layers with $4 \times 4$ filter, \# of input channels: 24, \# of output channels: 12, stride $= 2$, \\ padding $= 1$
        \item ReLU activation layer
        \item Transposed-Convolution layers with $4 \times 4$ filter, \# of input channels: 12, \# of output channels: 3, stride $= 2$, \\ padding $= 1$
        \item Sigmoid activation layer
    \end{enumerate}
\end{itemize}

\begin{table*}[ht]
\centering
{\footnotesize 
\begin{tabular}{c|cc|cccc} \toprule
Defense & Clean (Last) & Clean (Best) & FGSM & PGD$^{20}$  & PGD$^{40}$ & AutoAttack\\  \midrule \midrule
Clean & 90.72 $\pm$ 0.10  &  90.86 $\pm$ 0.10 &  5.19 $\pm$ 1.03 &  0.0 $\pm$ 0.0  &  0.0 $\pm$ 0.0 &  0.0 $\pm$ 0.0 \\
\midrule
AT &  81.53 $\pm$ 0.21  &  81.76 $\pm$ 0.22 &  54.11 $\pm$ 0.37 &  37.06 $\pm$ 0.27  &  36.73 $\pm$ 0.20 &  35.89 $\pm$ 0.19 \\
\midrule
TART (64) & 83.13 $\pm$ 0.33  &  83.40 $\pm$ 0.19 &  54.69 $\pm$ 0.16 &  37.21 $\pm$ 0.14  &  36.83 $\pm$ 0.12 &  35.85 $\pm$ 0.01 \\
\midrule
TART (128) & \textbf{83.47 $\pm$ 0.22}  &  \textbf{83.73 $\pm$ 0.15} &  \textbf{55.27 $\pm$ 0.25} &  \textbf{37.72 $\pm$ 0.03}  &  \textbf{37.41 $\pm$ 0.13} &  \textbf{36.58 $\pm$ 0.08} \\
\bottomrule
\end{tabular}}
\caption{Test accuracies (\%) on CIFAR-10 with WRN-32-10. The performances over three trials are reported along with their standard deviation.}
\label{tab:autoencoder}
\end{table*}

To investigate how the dimensionality of the latent space affects performance, we conduct experiments on CIFAR-10 using two variants of an autoencoder: one with a 64-dimensional latent space and another with a 128-dimensional latent space. For the 64-dimensional case, we also adjust the architecture accordingly by setting the number of output channels in the encoder’s ninth layer and the number of input channels in the decoder’s first layer to 64. All other experimental settings, including defense configurations, are identical to those described in Section~\ref{subsec:cifar}.

Table~\ref{tab:autoencoder} presents the performance comparison of clean training, AT, TART (64), and TART (128), where TART (64) and TART (128) use autoencoders with 64- and 128-dimensional latent spaces, respectively. We observe that even with a low-dimensional latent space (64), TART improves clean accuracy without sacrificing robustness. Moreover, TART (128) achieves even better performance, likely due to its ability to provide a more accurate estimation of the tangent space, resulting in more effective guidance during training.

\section{Ablation Study on Perturbation Bound Assignment}\label{app:epsilon_assignment}


In TART, the perturbation bound, $\epsilon$, is adaptively assigned to each data point to improve clean accuracy while maintaining model robustness. Our core approach, TART, employs a straightforward strategy: it allocates a maximum bound of $\epsilon_{\max}$ to the top 50\% of data points with the largest tangential components and a bound of $0$ to the bottom 50\%. The primary advantage of this simple, binary assignment is its significant computational efficiency. By using only two perturbation bounds, $0$ and $\epsilon_{\max}$, our method avoids the time-consuming process of regenerating adversarial examples, a factor that accounts for the majority of the training time in many adversarial training methods.

To validate this design choice, we conducted an ablation study on alternative assignment strategies. We considered two main variants:
\begin{itemize}
    \item TART-$\alpha$: A generalization of our core approach, which assigns a bound of $\epsilon_{\max}$ to the top $\alpha\%$ of data points and $0$ to the remaining $(1-\alpha)\%$.
    \item TART-prop: A proportional method that assigns the perturbation bound based on the magnitude of the tangential component (TC), given by the formula $\epsilon_i = \frac{TC_i}{\max_j(TC_j)} \cdot \epsilon_{\max}$.
\end{itemize}

Our experimental results, summarized in Table~\ref{tab:additional}, demonstrate that all three variants (TART (equivalent to TART-50), TART-75, and TART-prop) successfully improve clean accuracy without degrading robustness. While TART-prop also surpasses standard AT (equivalent to TART-100), its clean and robust performance lags behind TART. Furthermore, while TART-75 achieves similar or slightly better robustness than TART, its clean accuracy drops significantly. Given that TART-prop requires additional training time due to its non-binary assignment, our findings confirm that the simple assignment strategy of TART is highly effective while maintaining superior training efficiency.

\begin{table}[h]
\centering
\begin{tabular}{c|c|ccc} \toprule
Defense & Clean & FGSM & PGD$^{20}$  & AutoAttack\\  \midrule 
AT &  81.53  & 54.11 & 37.06 &  35.89  \\
TART & 83.47  &  55.27  &  37.72 & 36.58  \\
TART-75 &  82.14   & 54.91 &  37.99   &  36.79 \\
TART-prop & 83.38  &  54.86   & 37.30 & 36.08  \\
\bottomrule
\end{tabular}
\caption{Evaluation of Different Assignment Methods.}\label{tab:additional}
\end{table}

\end{document}